# Analysis of Word Embeddings using Fuzzy Clustering

Shahin Atakishiyev
*Department of Electrical and Computer Engineering*
*University of Alberta*
*Edmonton, AB, Canada*
shahin.atakishiyev@ualberta.ca

Marek Z. Reformat
*Department of Electrical and Computer Engineering*
*University of Alberta*
*Edmonton, AB, Canada*
marek.reformat@ualberta.ca

*Abstract* – In data dominated systems and applications, a concept of representing words in a numerical format has gained a lot of attention. There are a few approaches used to generate such a representation. An interesting issue that should be considered is the ability of such representations – called embeddings – to imitate human-based semantic similarity between words. In this study, we perform a fuzzy-based analysis of vector representations of words, i.e., word embeddings. We use two popular fuzzy clustering algorithms on count-based word embeddings, known as GloVe, of different dimensionality. Words from WordSim-353, called the gold standard, are represented as vectors and clustered. The results indicate that fuzzy clustering algorithms are very sensitive to high-dimensional data, and parameter tuning can dramatically change their performance. We show that by adjusting the value of the fuzzifier parameter, fuzzy clustering can be successfully applied to vectors of high – up to one hundred – dimensions. Additionally, we illustrate that fuzzy clustering allows to provide interesting results regarding membership of words to different clusters.

## I. INTRODUCTION

Word embeddings become an essential element of methods that focus on analysis and comparison of texts. One of the most popular embeddings is *GloVe* [10]. The embedding is obtained via analysis of word-word co-occurrences in a text corpus. A natural question is related to the ability of the embeddings to represent a human-based semantic similarity of words.

Data clustering is the process of grouping objects in a way that similarity between data points that belong to the same group (cluster) becomes as high as possible, while similarity between points from different groups gets as small as possible. It is an important task in analysis processes and has been successfully applied to pattern recognition [1] [2], image segmentation [3], fault diagnosis and search engines [4]. Fuzzy clustering which allows data points to belong to several numbers of clusters with different membership grades has been proved to have useful applications in many areas. Specifically, fuzzy C-Means (FCM) clustering and its augmented version – fuzzy Gustafson-Kessel (FGK) clustering are the most popular fuzzy clustering techniques. High-dimensional spaces often have a devastating effect on data clustering in terms of performance and quality; this issue is regarded as the *curse of the dimensionality*.

Our study sheds some light on the comparative analysis of the above fuzzy clustering methods to observe if and how the results of fuzzy clustering change with the dimensionality of word embeddings. Additionally, we illustrate 'usefulness' of fuzzy clustering via analysis of degrees of belongings of words to different clusters.

The paper is structured as follows: Section II contains the description of fuzzy C-Means (FCM) and fuzzy Gustafson-Kessel (FGK) algorithms. In Section III, we provide an overview of the validity indices applied for fuzzy clustering processes. The theoretical background behind the construction process of *GloVe* embeddings is covered in Section IV. Section V outlines the methodology, while Section VI shows our experimental results. Finally, the obtained conclusions are presented in Section VII.

## II. FUZZY CLUSTERING

Zadeh's fuzzy sets theory [5] has triggered a number of studies focused on the application of theoretical and empirical concepts of fuzzy logic to data clustering. In contrast to hard clustering techniques, where one point is assigned exactly to only one cluster, fuzzy clustering allows data points to pertain to several clusters with different grades of membership.

We have analyzed the behavior of FCM and FGK in clustering of vector representations of words in different dimensions. The details of these clustering methods have been described in the following sections.

*A. Fuzzy C-means clustering*

Fuzzy C-means algorithm was introduced by Bezdek [6] in 1981. It allows an observation to belong to multiple clusters with varying grades of membership. Having D as the number of data points, $N$ as the number of clusters, $m$ as the fuzzifier parameter, $x_i$ as the $i$-th data point, $c_i$ as the center of the $j$-th cluster, $\mu_{ij}$ as the membership degree of $x_i$ for the $j$-th cluster, FCM aims to minimize

$$J = \sum_{i=1}^{D}\sum_{j=1}^{N} \mu_{ij}^{m} ||x_i - c_j||^2 \quad (1)$$

The FCM clustering proceeds in the following way:

1. Cluster membership values $\mu_{ij}$ and initial cluster centers are initialized randomly.
2. Cluster centers are computed according to the formula:

$$c_j = \frac{\sum_{i=1}^{D} \mu_{ij}^{m} x_i}{\sum_{i=1}^{D} \mu_{ij}^{m}} \quad (2)$$

3. Membership grades ($\mu_{ij}$) are updated in the following way:

$$\mu_{ij} = \frac{1}{\sum_{1}^{N}\left(\frac{||x_i-c_j||}{||x_i-c_k||}\right)^{\frac{2}{m-1}}}, \mu_{ij} \in [0,1] \text{ and } \sum_{i=1}^{c}\mu_{ij} = 1 \quad (3)$$

4. The objective function $J$ is calculated
5. The steps 2,3,4 are repeated until the value of the objective function gets less than a specified threshold.

Fuzzy C-means (FCM) has many useful applications in medical image analysis, pattern recognition, and software quality prediction [6,7], to name just a few. The most important factors affecting the performance of this algorithm are the fuzzifier parameter *m*, the size and the dimensionality of data. The performance analysis of the algorithm for high-dimensional clustering will be discussed in Section VII in detail.

*B. Fuzzy Gustafson-Kessel clustering*

Fuzzy Gustafson-Kessel (FGK) extends FCM by introducing an adaptive distance norm that allows the algorithm to identify clusters with different geometrical shapes [8]. The distance metric is defined in the following way:

$$D_{GK}^2 = (x_k - v_i)^T A_i (x_k - v_i) \quad (4)$$

where $A_i$ itself is computed from a fuzzy covariance matrix of each cluster:

$$A_i = (\rho_i |C_i|)^{1/d} C_i^{-1}, \quad (5)$$

$$C_i = \frac{\sum_{k=1}^{N} \mu_{ik}^{m} (x_k - v_i)^T (x_k - v_i)}{\sum_{k=1}^{N} \mu_{ik}^{m}} \quad (6)$$

Here the parameter $\rho_i$ is the constrained form of the determinant of $A_i$, and $v_i$ is the center of a specified cluster *i*.

$$|A_i| = \rho_i, \ \rho_i > 0, \forall i \quad (7)$$

Enabling the matrix $A_i$ to change with fixed determinant serves to optimize the shape of clusters by keeping the cluster's volume constant [8]. Gustafson-Kessel clustering minimizes the following criterion:

$$J = \sum_{i=1}^{c}\sum_{k=1}^{N} \mu_{ik}^{m} D_{GK}^2 = \sum_{i=1}^{c}\sum_{i=1}^{N} \mu_{ik}^{m}(x_k - v_i)^T A_i(x_k - v_i) \quad (8)$$

Like FCM, this optimization is also subject to the following constraints:

$$\mu_{ik} \in [0,1], \forall i, k \text{ and } \sum_{i=1}^{c} \mu_{ik} = 1, \forall k \quad (9)$$

We see that the computation of the FGK algorithm is more convoluted than FCM clustering.

III VALIDITY INDICES

There are several validity indices to analyze the performance of the fuzzy clustering algorithms. One of them was proposed by Bezdek [6]. It is called fuzzy *partition coefficient (FPC)*. This index is calculated as follows:

$$FPC = \frac{1}{N}\sum_{k=1}^{N}\sum_{i=1}^{c} \mu_{ik}^2 \quad (10)$$

FPC changes between [0,1] range and the maximum value indicates the best clustering quality. Another popular index to measure fuzzy clustering quality was proposed by Xie and Beni (XB) in 1991 [9]. It focuses on two properties: cluster compactness and separation:

$$XB = \sum_{i=1}^{c} \frac{\sum_{k=1}^{N}(\mu_{ik})^m ||x_k - v_i||^2}{N min_{ik} ||x_k - v_i||^2} \quad (11)$$

The numerator part shows the strength of the compactness of fuzzy clustering, and the denominator shows the strength of separation between those fuzzy clusters. If a range of clusters { $k_1, k_2 \ldots k_i$} is taken, the $k_i$ minimizing this index will be the optimal number of clusters for the dataset.

IV. GloVe VECTORS

One of the most known unsupervised learning algorithm to produce vector representations of words is *GloVe*. It is based on a word-word co-occurrence in text corpora. The term stands for **Glo**bal **Ve**ctors as the representation is able to capture global corpus statistics.

*A. Overview of GloVe*

Let us start with defining a word-word co-occurrence counts matrix as *X*, where $X_{ij}$ is the number of times the word *j* exists in the context of the word *i*. Let use denote $X_i = \sum_k X_{ik}$ the number of times any word appears in the context of the word *i*. Lastly, let $P_{ij} = P(j|i) = X_{ij}/X_i$ become the probability in which word *j* exists in the context of the word *i*. Using a simple example, we demonstrate how aspects of meanings can be extracted from the word-word co-occurrences statistics. Pennington et al. show this with good examples. Assuming we have text corpora related to thermodynamics and we may take words $i = ice$ and $j = steam$. We investigate the relationship of these words by learning from the co-occurrence probabilities with other sample words. For instance, taking the word *ice* and word *t=solid*, we can expect that $P_{it}/P_{jt}$ will be large. Likewise, if we select words *t* that are related to *steam* but not to *ice* such that *t=gas,* then we expect that the value of ratio should be small. Global vectors try to leverage a series of functions called *F* that represents those ratios [10] [11]. These F functions for the ratio of $P_{it}/P_{jt}$ depend on words *i*, *j*, *t* to reflect the vector space models with linear structures:

$$F(w_i, w_j, w_t^{\sim}) = \frac{P_{it}}{P_{jt}} \quad (12)$$

where $w \in R$ are real word vectors and $w_t^{\sim} \in R$ are context word vectors. In order to attain the symmetry, we require F to be a homomorphism and eventually express Eq. (12) as:

$$\frac{F(w_i^T w_t^{\sim})}{F(w_j^T w_t^{\sim})} = \frac{P_{it}}{P_{jt}} \quad (13)$$

Adding bias terms for the $b_i$ and $b_t^{\sim}$ for the vectors $w_i$ and $w_t^{\sim}$ and expressing F=*exp*,

$$w_i^T w_t^{\sim} + b_i + b_t^{\sim} = \log(X_{it}) \quad (14)$$

One disadvantage of the Eq. (14) is that the logarithm diverges when its argument becomes 0. An optimal solution to deal with this problem is to represent the right side as $\log(1 + X_{it})$ where it preserves the sparsity of *X* and avoid the divergence. Based on the above method, the objective function for Glove which combines a least squares regression model with the weight function $f(X_{ij})$ is defined in the following way:

$$J = \sum_{i,j=1}^{V} f(X_{ij})(w_i^T w_j^{\sim} + b_i + b_j^{\sim} - \log X_{ij})^2 \quad (15)$$

Here V is the size of the vocabulary and $X_{ij}$ shows the number of times the word *j* exists in the context of the word *i*.

### B. Training GloVe and Data Corpora

The objective to train *GloVe* model is to find appropriate vectors that minimize the objective function in Eq. (15). As standard gradient descent algorithm heavily depends on the same learning rate, it does not become helpful to find errors and update them properly. Adaptive gradient algorithm (AdaGrad) has been proposed to solve the problem which adaptively assigns different learning rates to each of parameters [10] [11]. After training, the model produces two sets of vectors: *W* and *W* ˜. When *X* is symmetric, the generated word vectors intrinsically perform equally and can become different only owing to random initializations. The authors show the best way to handle with these two vectors is to sum and assign the sum vector as a unique representation of the word:

$$W_{final} = W + W^{\sim} \quad (16)$$

Summing two sets of vectors into one effectively reflects words in the embeddings space. The authors have built the vocabulary of most frequent 400,000 words, and made them publicly available with 50,100, 200 and 300 dimensions, under Public Domain Dedication and License[1]. The source of the training data can be seen at [10].

### V. METHODOLOGY

Before we provide a description of the obtained results and their analysis, we briefly describe a set of words that has been clustered, as well as a procedure used for determining values of some clustering parameters.

### A. Gold Standard for Similarity Evaluation

A set of words represented by word embeddings that we cluster has been constructed using words of WordSim-353 [12] dataset. This dataset contains semantic similarity scores of 353 word pairs and contains 437 different words. These pairs have been merged from 153 pairs scored by 13 humans and 200 pairs scored by 16 humans. The semantic similarity scores for the pairs vary in the range of [0-10]. For example, the similarity measures for the words *journey* and *voyage*, *computer* and *internet*, and *media* and *gain* are 9.29, 7.58 and 2.88, respectively. Many researchers have referred to WordSim-353 as a *gold standard* for different word similarity tasks. We have extracted vector representations of those gold words from 50, 100, 200 and 300-dimensional versions of *GloVe*, and used them for clustering and further analysis.

### B. Clustering Parameters

One of the important parameters of a clustering process is a number of clusters. The nature of unsupervised learning means that this number needs to be set a priori. In our experiments, we use a t-SNE [13] visualization of WordSim-353 words (Fig. 1) to define a minimum number of clusters. The range of the number of clusters is determined in the following way.

Lower Boundary. We use a simple visualization of words based on t-SNE, Fig. 1. Based on a visual inspection, we have identified the most obvious groups of words. As you can see, there are ten locations characterized by a higher concentration of words. Therefore, we use ten as our lower boundary for the number of clusters.

Upper Boundary. There are 437 words in the dataset we use in clustering experiments. We have anticipated that a larger number of clusters would provide better performance in the sense of clustering performance measures. However, we would like to avoid creating too small clusters – smaller cluster would be counterintuitive to our need for observing pairs of words in clusters (Section VI.B). Therefore, we have established the acceptable smaller size, on average, of a cluster to around 10. That would lead to a maximum of 50 cluster – and this becomes our upper boundary for the number of clusters.

---

[1]Pre-trained 400,000 GloVe vectors available in: https://nlp.stanford.edu/projects/glove/

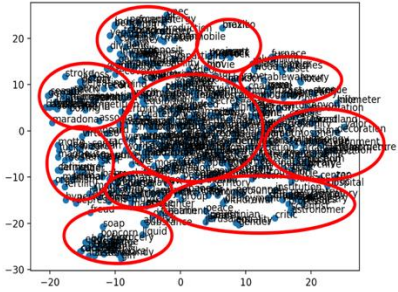

Fig 1. t-SNE Visualization of Word Vectors

## VI. EMPIRICAL RESULTS

The results of fuzzy clustering of WordSim-353 words represented with *GloVe* embeddings are shown in Tables 1, 2, and 3.

### A. Quantitative Analysis

It is well-known that fuzzy clustering shows some problems in the case of clustering high dimensional data [14]. Winkler et al. have shown that performance of fuzzy clustering dramatically changes with the fuzzifier parameter: when using *m=2*, the majority of prototypes go into the center of the gravity of the whole dataset. Therefore, we neither acquire the expected number of clusters nor sensible clustering results. Adjusting the *fuzzifier* around 1 such as 1.1 substantially improves the performance of the clustering. As a result, we obtain high-quality groupings of data points until some dimensions.

Based on that, we have set the value of the *fuzzifier* parameter to *m=1.1* for our experiments. As we can see in the tables, clustering of words with 200-dim embeddings (Table 3) results with the performance measures values that start to look quite unreasonable (Xie-Beni index), while values of the index FPC become quite small. Due to such a situation, we consider for further analysis (next subsection) clusters obtained using FCM and FGK with 50-dim *GloVe* embeddings, and only clusters obtained using FCM for 100-dim embedding.

Table 1. Clustering results with 50-dimensional *GloVe* Embeddings

| # of clusters | 10 | 15 | 20 | 25 | 30 | 40 | 50 |
|---|---|---|---|---|---|---|---|
| Fuzzy C-Means | | | | | | | |
| Xie-Beni index | 0.0054 | 0.0053 | 0.0053 | 0.0056 | 0.0053 | 0.0047 | **0.0041** |
| FPC | 0.7246 | 0.7242 | 0.7476 | 0.8021 | 0.8227 | 0.8683 | **0.8878** |
| Fuzzy Gustafson-Kessel | | | | | | | |
| Xie-Beni index | 16.48 | 13.43 | 10.12 | 13.43 | 11.58 | 9.53 | **8.00** |
| FPC | 0.9999 | 0.9878 | 0.9863 | 0.9875 | 0.9874 | 0.9879 | **0.9892** |

Table 2. Clustering results with 100-dimensional *GloVe* Embeddings

| # of clusters | 10 | 15 | 20 | 25 | 30 | 40 | 50 |
|---|---|---|---|---|---|---|---|
| Fuzzy C-Means | | | | | | | |
| Xie-Beni index | 0.0096 | 0.0122 | 0.0092 | 0.0081 | 0.0108 | 0.0080 | **0.0076** |
| FPC | 0.5917 | 0.5987 | 0.6471 | 0.6980 | 0.7210 | 0.7817 | **0.8322** |
| Fuzzy Gustafson-Kessel | | | | | | | |
| Xie-Beni index* | 30.03 | 25.40 | 21.56 | 20.58 | 15.11 | 12.65 | 10.87 |
| FPC | **0.9817** | 0.9783 | 0.9708 | 0.9747 | 0.9713 | 0.9648 | 0.9701 |

*numbers in gray represent unacceptable values

Table 3. Clustering results with 200-dimensional *GloVe* Embeddings

| # of clusters | 10 | 15 | 20 | 25 | 30 | 40 | 50 |
|---|---|---|---|---|---|---|---|
| Fuzzy C-Means | | | | | | | |
| Xie-Beni index* | 30997.7 | 8685.2 | 10542791 | 156571.2 | 641690.1 | 572974.6 | 30602.8 |
| FPC | 0.2612 | 0.2896 | 0.3470 | 0.3638 | 0.4935 | 0.5714 | **0.5907** |
| Fuzzy Gustafson-Kessel | | | | | | | |
| Xie-Beni index* | 696.62 | 731.98 | 706.72 | 728.19 | 782.20 | 769.02 | 749.63 |
| FPC | **0.4154** | 0.2978 | 0.2276 | 0.1842 | 0.1568 | 0.1194 | 0.0951 |

*numbers in gray represent unacceptable values

Table 4: GloVe Embeddings: number of word pairs found in clusters

| GloVe dimensionality: | 50 | 100 |
|---|---|---|
| Fuzzy C-Means | 42.40 +/-3.03 | 47.50 +/- 2.42 |
| Fuzzy Gustafson-Kessel | 9 | 5 |

When the fuzzifier parameter *m* equals *1*, the results become crisp. So, one question can naturally be raised: can a clustering with *m=1.1* be still considered as a fuzzy clustering? For this purpose, we have analyzed the fuzzy membership matrix to see the distribution of the memberships. Among all words we have clustered, there are 25 words that have at the maximum membership value of 0.75 to a single cluster for 50-dim embeddings, and 52 such words for 100-dim embedding. We analyze the obtained clusters in Section VI.C.

*B. Qualitative Analysis*

The results presented in Section VI.A describe clusters from the point of view of their quality as measured by the performance indexes. However, these indexes do not show how well the clusters and clustering techniques group semantically similar words. For this purpose, we propose another way of determining the quality of clusters. The proposed approach is done from the point of view of grouping similar – according to humans – words.

The first step in the proposed approach has been to identify a set of pairs of words that are *highly* similar. Here, we use *the gold standard*, i.e., the set WordSim-353. We have assumed that the similarity value of 0.75 could be considered as a reasonable and practical level of treating words as *highly* similar. As a result, we have obtained 93 pairs of words.

The second step of the approach is to determine the number of pairs that are present in the same cluster. Moreover, we look at the distribution of pairs among clusters, i.e., we have identified clusters with zero, one, two and so on a number of pairs. The results of ten experiments are presented in Table 4. It shows an average number of word pairs found in the same clusters. As we can see, FCM is the best performing clustering technique. Also, an increase in the dimensionality of word representation leads to better results. This observation is statistically significant with the value of $p < 0.01$. In the case of FGK, the obtained numbers of pairs in clusters are very low. Also, ten experiments have led to the same numbers: 9 for 50-dim and 5 for 100-dim word representations.

*C. Analysis of Fuzzy Clusters*

The usage of fuzzy clustering means that we obtain fuzzy clusters with data point – words in our case – that belong to a cluster to a degree. Therefore, let us analyze some examples of allocating – to a degree – words to different clusters. We show two cases: 1) one that illustrates how two words are 'shared' between four clusters; and 2) other one that demonstrates how two words belong to different degrees to two clusters.

The first case is presented in Fig. 2. It shows two words **earning** and **marathon** that belong to different degrees to four different clusters. The content of these clusters is:

**A**: {'card', 'listing', 'category', 'viewer', 'ticket', 'cd'};

**B**: {'wealth', 'entity', 'image', 'recognition', 'confidence', 'gain', 'importance', 'prominence', 'credibility'};

**C**: {'string', 'record', 'number', 'hundred', 'row', 'place', 'five', **'earning'**, **'marathon'**, 'series', 'start', 'year', 'day', 'summer', 'performance', 'seven'}; and

**D**: {'football', 'soccer', 'basketball', 'tennis', 'star', 'cup', 'medal', 'competition', 'baseball', 'season', 'game', 'team', 'boxing', 'championship', 'tournament', 'world'}.

The word **earning** is a member of three different clusters: **A** to a degree of 0.10, **B** to a degree of 0.15, and **C** to a degree of 0.50. It seems that its main cluster is *C*. While the word **marathon** belongs to **C** to a degree of 0.40, and to **D** to a degree of 0.30. If we look at the words from each cluster and the two considered words, we can easily see that their different degrees of membership to clusters are fully justified. **Earning** 'makes' sense to have some relationship with clusters A and B, while **marathon** could 'easily' be a member of cluster D.

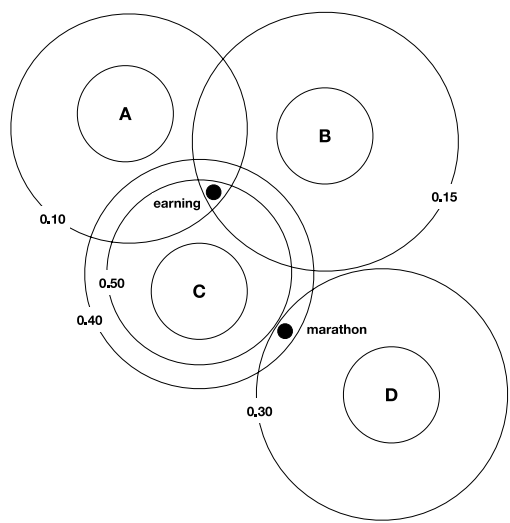

Fig 2. Visualization of words: **earning** and **marathon** that belong to multiple clusters

The second example shows two words that belong to the same two clusters to different degrees, Fig. 3. Both clusters consist of:

- **X**: {'space', 'example', 'object', 'weapon', 'surface', **'activity'**, 'type', 'combination', 'proximity', 'cell', 'size', 'observation'}, and
- **Y**: {'fear', 'mind', **'atmosphere'**, 'reason', 'problem', 'kind', 'situation', 'direction', 'lesson', 'focus', 'change', 'attitude', 'approach', 'practice', 'experience'}

The word **activity** belongs to **X** to a degree of 0.55 while to **Y** to a degree of 0.15, while **atmosphere** has a membership value of 0.35 to **X**, and 0.40 to **Y**. Once again, we see that different degrees of membership to the clusters are fully justified.

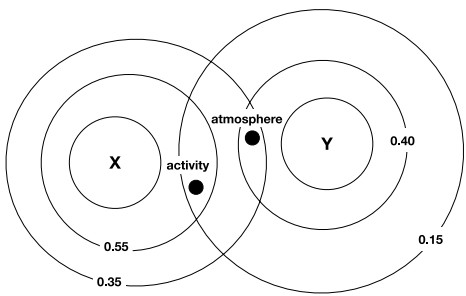

Fig 3. Visualization of words: **activity** and **atmosphere** that belong to two clusters

## VII. CONCLUSIONS

In this study, we have examined the fuzzy clustering analysis of word embeddings. We analyzed the performance of fuzzy C-means and fuzzy Gustafson-Kessel algorithms on the set of words from the WordSim-353 dataset represented using *GloVe* embeddings. Based on the obtained results, fuzzy clustering algorithms were proved to be very sensitive for high-dimensional data. Fuzzy C-means with a fuzzifier parameter *m=1.1* has provided sensible word clustering results for up to 100-dimensional word embeddings. However, in larger dimensions, it fails to produce both the expected number of clusters and plausible word clustering results. The experimental results proved that fuzzy Gustafson-Kessel clustering technique, on the other hand, should be avoided in high-dimensional spaces. Even for the case of 50-dimensional data, a very poor performance has been observed.

Additionally, we have shown that using fuzzy clustering with a small value of a fuzzifier parameter (*m=1.1*) still provides an interesting and fully justified variation in the degrees of membership of words to different clusters.

ACKNOWLEDGEMENTS

The authors express their gratitude to the Ministry of Education of the Republic of Azerbaijan for funding this research under the "State Program on Education of Azerbaijani Youth Abroad in the Years of 2007-2015" program.